\documentclass[letterpaper]{article} 
\usepackage{aaai24}  
\usepackage{times}  
\usepackage{helvet}  
\usepackage{courier}  
\usepackage[hyphens]{url}  
\usepackage{graphicx} 
\usepackage{subfig}
\usepackage{booktabs}
\usepackage{multirow}
\usepackage{amsmath}
\usepackage{physics}
\usepackage{adjustbox}
\usepackage{xcolor}
\usepackage{float}
\usepackage{array}

\usepackage{natbib}  
\usepackage{caption} 
\usepackage{algorithm}
\usepackage{algorithmic}

\usepackage{newfloat}
\usepackage{listings}
\DeclareCaptionStyle{ruled}{labelfont=normalfont,labelsep=colon,strut=off} 
\floatstyle{ruled}
\newfloat{listing}{tb}{lst}{}
\floatname{listing}{Listing}
\title{TC-LIF: A Two-Compartment Spiking Neuron Model\\ for Long-Term Sequential Modelling}
\author{
    Shimin Zhang\textsuperscript{\rm 1}\equalcontrib,
    Qu Yang\textsuperscript{\rm 2}\equalcontrib,
    Chenxiang Ma\textsuperscript{\rm 1},
    Jibin Wu\textsuperscript{\rm 1}\thanks{Corresponding Author},
    Haizhou Li\textsuperscript{\rm 3, \rm 2}
    Kay Chen Tan\textsuperscript{\rm 1}
}
\affiliations{
    \textsuperscript{\rm 1}The Hong Kong Polytechnic University, Hong Kong SAR, China\\
    \textsuperscript{\rm 2}National University of Singapore, Singapore\\
    \textsuperscript{\rm 3}The Chinese University of Hong Kong, Shenzhen (CUHK-Shenzhen), Guangdong, China\\
    shi-min.zhang@connect.polyu.hk, quyang@u.nus.edu, chenxiang.ma@connect.polyu.hk, jibin.wu@polyu.edu.hk, haizhou.li@u.nus.edu, kctan@polyu.edu.hk
}

\usepackage{bibentry}

\begin{document}

\maketitle

\begin{abstract}
The identification of sensory cues associated with potential opportunities and dangers is frequently complicated by unrelated events that separate useful cues by long delays. As a result, it remains a challenging task for state-of-the-art spiking neural networks (SNNs) to establish long-term temporal dependency between distant cues. To address this challenge, we propose a novel biologically inspired \textbf{T}wo-\textbf{C}ompartment \textbf{L}eaky \textbf{I}ntegrate-and-\textbf{F}ire spiking neuron model, dubbed TC-LIF. The proposed model incorporates carefully designed somatic and dendritic compartments that are tailored to facilitate learning long-term temporal dependencies. Furthermore, a theoretical analysis is provided to validate the effectiveness of TC-LIF in propagating error gradients over an extended temporal duration. Our experimental results, on a diverse range of temporal classification tasks, demonstrate superior temporal classification capability, rapid training convergence, and high energy efficiency of the proposed TC-LIF model. Therefore, this work opens up a myriad of opportunities for solving challenging temporal processing tasks on emerging neuromorphic computing systems. Our code is publicly available at \textit{https://github.com/ZhangShimin1/TC-LIF}.
\end{abstract}

\section{Introduction}
\label{sec: intro}
Spiking neural networks (SNNs) have attracted significant attention recently owing to their biological plausibility and potential for energy-efficient neural computation \cite{maass1997networks, pfeiffer2018deep}. The fundamental computing units of SNNs, known as spiking neurons, aim to emulate the rich neuronal dynamics observed in biological neurons, which facilitate the encoding, processing, and storage of spatio-temporal patterns~\cite{gerstner2014neuronal}. Furthermore, spiking neurons communicate with each other via discrete spikes, and such event-driven operation leads to ultra-low-power neural computation \cite{davies2018loihi,pei2019towards}, achieving efficient learning~\cite{Xinyin23DeepCache,Gongfan23DepGraph,Songhua22DD}.

In practice, single-compartment spiking neurons models have been widely adopted for large-scale brain simulations and neuromorphic computing, instances include Leaky Integrate-and-Fire (LIF) Model \cite{abbott2005model}, Izhikevich Model \cite{izhikevich2003simple}, and Adaptive Exponential Integrate-and-Fire (AdEx) Model \cite{brette2005adaptive}. These single-compartment models abstract the biological neuron as a single electrical circuit, preserving the essential neuronal dynamics of biological neurons while ignoring the complex geometrical structure of dendrites as well as their interactions with the soma. This degree of abstraction significantly reduces the modeling effort, making them more feasible to study the behavior of large-scale biological neural networks and perform complex pattern recognition tasks on neuromorphic computing systems \cite{wu2021tandem, 10347028, ma2022deep}.

While single-compartment spiking neuron models have demonstrated promising results on pattern recognition tasks with limited temporal context \cite{tavanaei2019deep, zhang2021rectified, wu2021progressive, yang2022training, yao2022glif}, their ability to solve tasks that require long-term temporal dependencies remains constrained. 
This issue primarily arises from the difficulty of performing long-term temporal credit assignment (TCA) in SNNs. The TCA involves the identification of input spikes that contribute to future rewards or penalties, and subsequently strengthen or weaken their respective connections. Given the discrete and sequential nature of spikes, pinpointing the exact moments or sequences that led to a prediction error becomes challenging. This issue deteriorates for long sequences, where the influence of early spikes on later predictions is more challenging to trace. Hence, addressing the TCA problem is crucial for enhancing the sequential modeling capabilities of SNNs.

Broadly speaking, there are two research directions have been pursued to address the TCA problem in SNNs. The first direction draws inspiration from the recent success of attention models within deep learning. These methods integrate the self-attention mechanism into SNNs to enable the direct modeling of temporal relationships between different time steps~\cite{qin2023attention, yao2021temporal}. However, the self-attention mechanism is computationally expensive to operate, and it is unable to operate in real-time. Furthermore, self-attention is not directly compatible with mainstream neuromorphic chips, therefore, it cannot take advantage of the energy efficiency offered by these chips. 

\begin{figure*}[htb]
\centering
\subfloat[]{
\begin{adjustbox}{raise=0.1cm}
    \includegraphics[width=55mm]{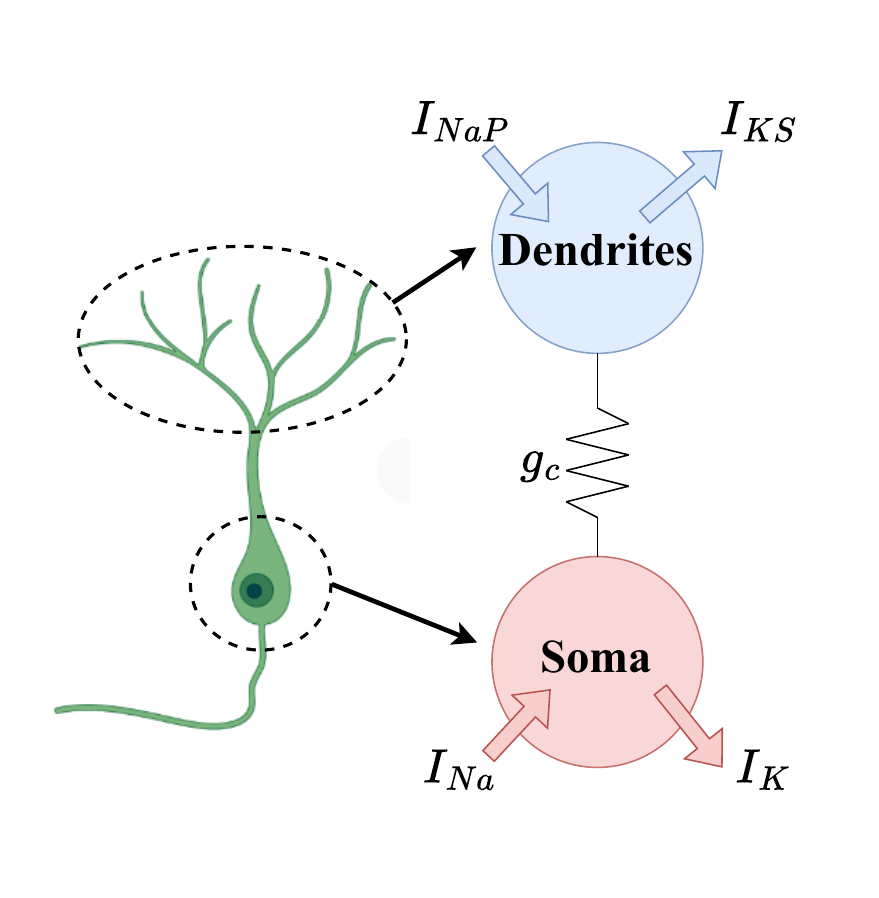}
\end{adjustbox}
} \quad
\subfloat[]{
\begin{adjustbox}{raise=0.05cm}
    \includegraphics[width=50mm]{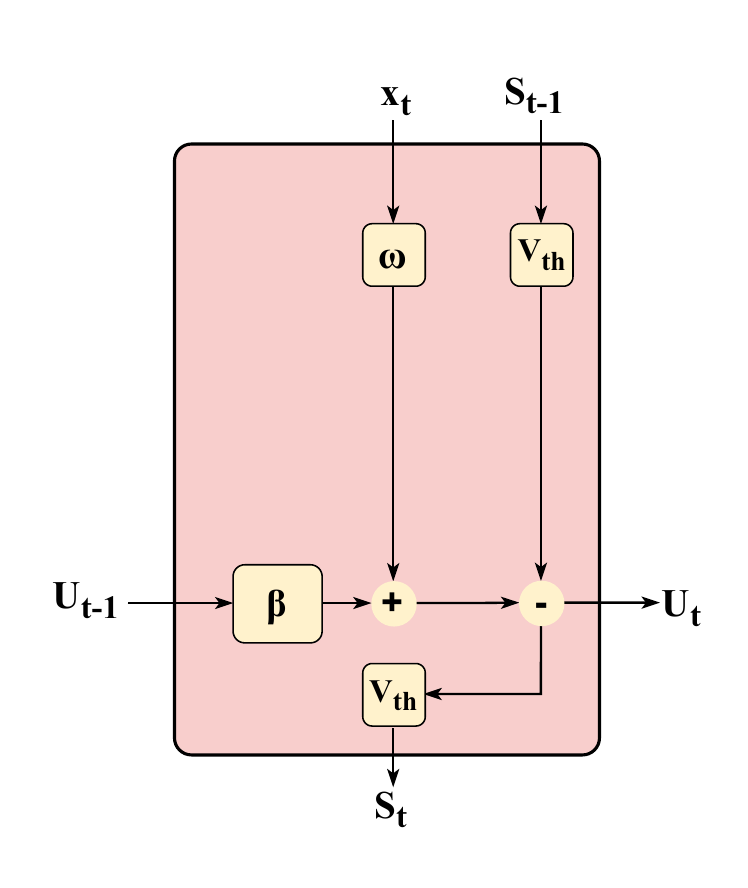}
\end{adjustbox}
} \quad \quad 
\subfloat[]{\includegraphics[width=51mm]{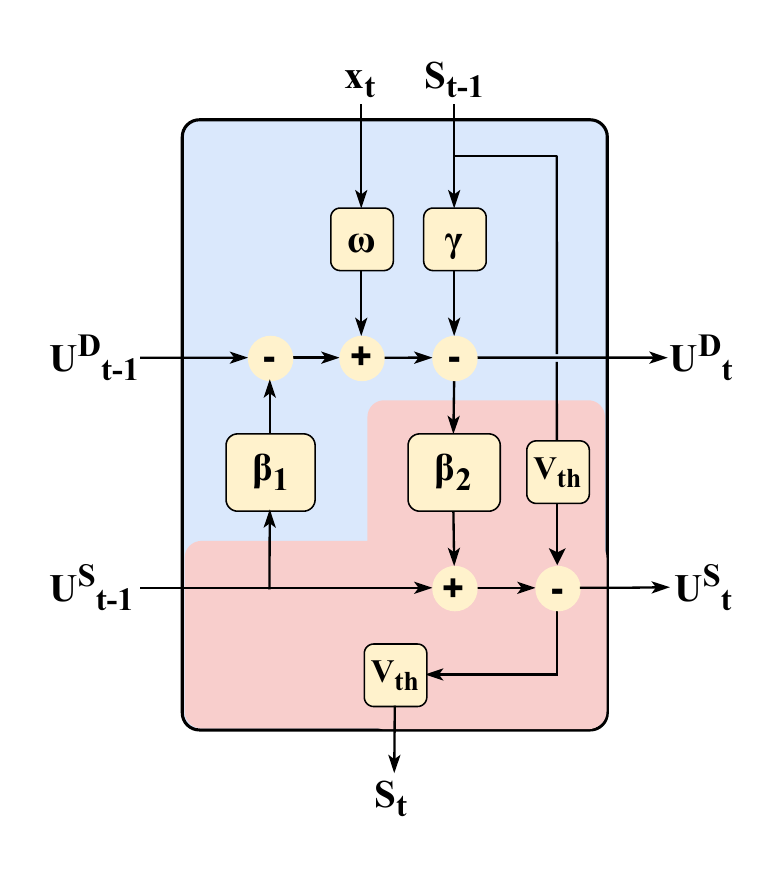}} 
\caption{Illustration of (a) the structure of a two-compartment Pinsky-Rinzel pyramidal neuron, and the internal operations of (b) LIF model as well as the proposed (c) TC-LIF model. Note that the proposed TC-LIF model in (c) differentiates dendirtic and somatic compartments by highlighting the regions in blue and red, respectively. In contrast, the LIF model in (b) solely incorporates the dynamics of the somatic compartment, without accounting for the dendritic compartment. }
\label{fig: inter_ops}
\end{figure*}

The latter research direction primarily centers around the idea of adaptive spiking neuron models. Notable efforts include the Long Short-Term Memory Spiking Neural Network (LSNN) \cite{bellec2018long}. LSNN introduces an adaptive firing threshold mechanism into LIF neurons, whereby the neuronal firing threshold increases following each firing event and slowly decays back to the resting-state firing threshold. This elevation in the firing threshold serves as a means of information storage, particularly when combined with a slow decay rate, effectively facilitating long-term TCA \cite{bellec2020solution}. Additional studies have proposed the utilization of learnable time constants \cite{yin2020effective, yin2021accurate} or dual time constants \cite{shaban2021adaptive} for the adaptive firing threshold, such that multi-scale temporal information can be retained and leveraged to perform TCA. However, these studies have predominantly focused on enhancing the neuronal firing threshold, which, as a simple neuronal component, possesses inherent restrictions in terms of information storage capacity. Consequently, these models face inherent limitations in their capabilities to address the TCA problem.

Multi-compartment neuron models have been the subject of extensive research in the field of neuroscience~\cite{rall1964theoretical,pinsky1994intrinsic}.
These models aim to faithfully model the complex geometric structure of dendrites, along with the interactions between dendritic and somatic compartments. As a result, multi-compartment models provide a more accurate representation of the complex neuronal dynamics observed in biological neurons, facilitating information interaction across various temporal scales \cite{stuart2015dendritic}. 
Consequently, they present a promising avenue for addressing the challenge of long-term TCA. While incorporating more compartments offers additional benefits of expanded memory capacity, the increased model complexity may hinder their practical use, especially for solving complex pattern recognition tasks.

In this paper, we derive a generalized two-compartment neuron model as depicted in Figure \ref{fig: inter_ops}(a). This neuron model provides an ideal reflection of the minimal geometry of the biological neuron while preserving the essential features of more complicated multi-compartment models \cite{lin2017dynamical}. Based on this, we further proposed a two-compartment spiking neuron model, called TC-LIF (\textbf{T}wo-\textbf{C}ompartment \textbf{L}eaky \textbf{I}ntegrate-and-\textbf{F}ire), which is tailored to address the long-term TCA problem. The main contributions of our work are summarized as follows:
\begin{itemize}
\item We propose a brain-inspired two-compartment spiking neuron model, dubbed TC-LIF, which has been carefully designed to facilitate long-term sequential modeling.
\item We provide theoretical and experimental analysis to validate the effectiveness of the proposed TC-LIF model in achieving successful long-term TCA. 
\item Our experimental results, on a broad range of temporal classification tasks, demonstrate superior sequential modeling capabilities of TC-LIF over single-compartment neuron models, including enhanced classification accuracy, rapid training convergence, and high energy efficiency. 
\end{itemize}

\section{Methodology}
\label{sec: methodology}
In this section, we first introduce a conventional single-compartment neuron model, specifically the LIF model. We delve into its inherent limitation in effectively learning long-term dependencies. Then, we present a generalized two-compartment spiking neuron model inspired by the well-known Pinsky-Rinzel pyramidal neurons \cite{pinsky1994intrinsic}. This set the stage for the development of our proposed TC-LIF model, meticulously tailored to address the long-term TCA problem. Furthermore, we provide a theoretical analysis to elucidate the mechanisms through which the TC-LIF model effectively facilitates long-term TCA.

\subsection{LIF Neurons Struggle to Perform Long-term TCA}
\label{sec: SNN}
In general, spiking neurons integrate synaptic inputs, transduced from the incoming spikes, into their membrane potentials. Once the accumulated membrane potential surpasses the firing threshold, an output spike will be generated and transmitted to subsequent neurons. The LIF neuron is the most ubiquitous and effective single-compartment spiking neuron model, which has been widely used for large-scale brain simulation and neuromorphic computing. The neuronal dynamics of a LIF neuron can be described by the following discrete-time formulations:
\begin{align}
\mathcal{U}[t] &= \beta\hspace{0.5mm}\mathcal{U}[t-1] - \mathcal{V}_{th}\mathcal{S}[t-1] + \mathcal{I}[t] \label{LIF_dynamics_whole} \\
\mathcal{I}[t] &= \sum_{i}{\omega_{i}x_{i}[t]}+b \label{Inj_I} \\
\mathcal{S}[t] &= \Theta(\mathcal{U}[t]-\mathcal{V}_{th})
\label{Surrogate_spike_cal}
\end{align}
where $\mathcal{U}[t]$ and $\mathcal{I}[t]$ represent the membrane potential and the input current of a neuron at time $t$, respectively. The term $\beta \equiv exp(-dt/\tau_{m})$ is the membrane decaying coefficient that ranged from (0, 1), in which $\tau_m$ is the membrane time constant and $dt$ is the simulation time step. $x_{i}$ is the output spike of input neuron $i$ from the previous layer, $\omega_{i}$ denotes the synaptic weight that connects input neuron $i$, and $b$ represents the bias term. An output spike will be generated once the membrane potential $\mathcal{U}[t]$ reaches the neuronal firing threshold $\mathcal{V}_{th}$ as per Eq. (\ref{Surrogate_spike_cal}).

The backpropagation-through-time (BPTT) algorithm, coupled with surrogate gradients, has been recently proposed as an effective approach to perform credit assignment in SNNs \cite{wu2018spatio,neftci2019surrogate}. While this approach demonstrates effectiveness in tasks that involve limited temporal context, it encounters limitations when confronted with tasks that necessitate long-term temporal dependencies. This is primarily attributed to the vanishing gradient problem, where the error gradients diminish during the backpropagation process. To further elaborate on this issue, let us consider the training of a SNN with the following objective function:
\begin{equation}
\label{SNN_obj}
\mathcal{L}(\hat{\mathcal{S}}, \mathcal{S})=\frac{1}{N}\sum_{n=1}^{N}{\mathcal{L}}(\hat{\mathcal{S}}_{n}, \mathcal{S}_{n})
\end{equation}
where $N$ is the number of training samples, ${\mathcal{L}}$ is the loss function, ${\mathcal{S}}_{n}$ is the network output, and $\hat{\mathcal{S}}_{n}$ is the training target. Following the BPTT algorithm, the gradient with respect to the weight $\omega$ can be calculated as follows:
\begin{equation}
\label{SNN_gradient}
\frac{\partial\mathcal{L}}{\partial\omega}=\sum_{t}^{T}\frac{\partial\mathcal{L}}{\partial\mathcal{S}[T]}\frac{\partial\mathcal{S}[T]}{\partial\mathcal{U}[T]}\frac{\partial\mathcal{U}[T]}{\partial\mathcal{U}[t]}\frac{\partial\mathcal{U}[t]}{\partial\omega}
\end{equation}
where for a LIF neuron with the membrane decaying rate of $\beta\in(0,1)$:
\begin{equation}
\label{SNN_partial_U}
\frac{\partial\mathcal{U}[T]}{\partial\mathcal{U}[t]}=\prod_{i=t+1}^{T}\frac{\partial\mathcal{U}[i]}{\partial\mathcal{U}[i-1]}=\beta^{(T-t)}
\end{equation}
It is obvious that as the time step $T$ increases, the impact of time step $t$ on its subsequent time step diminishes. This is because the membrane potential decay causes an exponential decay of early information. This problem becomes exacerbated when $t$ is considerably smaller than $T$, and the value of Eq. (\ref{SNN_partial_U}) tends to 0, thus leading to the vanishing gradient problem. Consequently, the existing single-compartment neuron models, such as the LIF model, face challenges in effectively propagating gradients to significantly earlier time steps. This poses a significant limitation in learning long-term dependencies, which motivates us to develop two-compartment neuron models that possess enhanced capabilities in facilitating long-term TCA. 

\subsection{A Generalized Two-Compartment Spiking Neuron}
\label{sec: PR_model}
The P-R pyramidal neurons are located in the CA3 region of the hippocampus, which plays an important role in memory storage and retrieval of animals \cite{pinsky1994intrinsic}. Researchers have simplified this neuron model as a two-compartment model that can simulate the interaction between somatic and dendritic compartments, as depicted in Figure \ref{fig: inter_ops}(a). Drawing upon the structure of the P-R model, we develop a generalized two-compartment spiking neuron model that defined as the following. The detailed derivations of this formulation are provided in Supplementary Materials. 
\begin{align}
\mathcal{U}^{D}[t] &= \alpha_1\hspace{0.5mm}\mathcal{U}^{D}[t-1] + \beta_1\hspace{0.5mm}\mathcal{U}^{S}[t-1] + \mathcal{I}[t] \label{PR_dendrite_general} \\
\mathcal{U}^{S}[t]&=\alpha_2\hspace{0.5mm}\mathcal{U}^{S}[t-1] + \beta_2\hspace{0.5mm}\mathcal{U}^{D}[t] - \mathcal{V}_{th}\mathcal{S}[t-1] \label{PR_soma_general} \\
\mathcal{S}[t]&=\Theta(\mathcal{U}^{S}[t]-\mathcal{V}_{th}) \label{PR_spike_cal}
\end{align}
where $\hspace{0.5mm}\mathcal{U}^{D}$ and $\hspace{0.5mm}\mathcal{U}^{S}$ represents the membrane potentials of the dendritic and the somatic compartments, respectively. 
$\alpha_1$ and $\alpha_2$ are respective membrane potential decaying coefficients for these two compartments.
Notably, the membrane potentials of these two compartments are not updated independently. Rather, they are coupled with each other through the second term in Eqs. (\ref{PR_dendrite_general}) and (\ref{PR_soma_general}), in which the coupling effects are controlled by the coefficients $\beta_1$ and $\beta_2$.
The interplay between these two compartments enhances the neuronal dynamics and, if properly designed, can resolve the vanishing gradient problem.

\subsection{TC-LIF Spiking Neuron Model}
\label{sec: TCSN}
Based on the generalized two-compartment spiking neuron model derived earlier, 
we propose a TC-LIF neuron model that has been carefully designed to facilitate long-term sequential modeling.
In comparison to the generalized two-compartment neuron model, we drop the membrane decaying factors $\alpha_1$ and $\alpha_2$ from both compartments. This modification aims to circumvent the rapid decay of memory that could cause unintended information loss. Moreover, to circumvent excess firing caused by persistent input accumulation, we set $\beta_1$ and $\beta_2$ to opposite signs. The dynamics of the proposed TC-LIF model are expressed as follows: 
\begin{align}
\mathcal{U}^{D}[t]&=\mathcal{U}^{D}[t-1] + \beta_1\hspace{0.5mm}\mathcal{U}^{S}[t-1] + \mathcal{I}[t] - \gamma\mathcal{S}[t-1] \label{TCME_dendrite} \\   
\mathcal{U}^{S}[t]&=\mathcal{U}^{S}[t-1] + \beta_2\hspace{0.5mm}\mathcal{U}^{D}[t] - \mathcal{V}_{th}\mathcal{S}[t-1] \label{TCME_soma} \\
\mathcal{S}[t]&=\Theta(\mathcal{U}^{S}[t]-\mathcal{V}_{th}) \label{TCME_spike_cal}
\end{align}
where the coefficients $\beta_1 \equiv -\sigma(c_1)$ and $\beta_2 \equiv \sigma(c_2)$ determine the interaction between these two compartments. Here, the sigmoid function $\sigma(\cdot)$ is utilized to ensure two coefficients are within the range of (-1, 0) and (0, 1), and the parameters $c_1$ and $c_2$ can be automatically adjusted during the training process. The effect of this design choice will be analyzed in detail soon.
The membrane potentials of both compartments are reset after the firing of the soma. Notably, the reset of the dendritic compartment is triggered by the backpropagating spike that is governed by a scaling factor $\gamma$. The internal operations of the TC-LIF model are depicted in Figure \ref{fig: inter_ops}(c), which exhibits richer internal dynamics in comparison to the LIF model that is shown in Figure \ref{fig: inter_ops}(b).

According to the above formulations, $\mathcal{U}^{S}$ is responsible for retaining short-term memory of dendritic inputs, which will be reset after neuron firing. In contrast, $\mathcal{U}^{D}$ serves as a long-term memory that retains the information about external inputs, which is only partially reset by the backpropagating spike from the soma. In this way, the multi-scale temporal information is effectively preserved in TC-LIF. To further demonstrate the superiority of TC-LIF in facilitating long-term TCA, we provide a theoretical analysis below to demonstrate how the TC-LIF model greatly alleviate the vanishing gradient problem. 

As discussed earlier, the primary cause of the gradient vanishing problem is attributed to the recursive computation of $\partial\mathcal{U}[T]/\partial\mathcal{U}[t]$.
This problem can, however, be effectively alleviated in the proposed TC-LIF model, whose partial derivative $\partial\mathcal{U}[T]/\partial\mathcal{U}[t]$ can be calculated as follows:
\begin{equation}
\label{TCME_adj_gradient_aggregation}
\frac{\partial\hspace{0.5mm}\mathcal{U}[T]}{\partial\hspace{0.5mm}\mathcal{U}[t]}=\prod_{j=t+1}^{T}\frac{\partial\hspace{0.5mm}\mathcal{U}[j]}{\partial\hspace{0.5mm}\mathcal{U}[j-1]}, \hspace{1.0mm}\mathcal{U}[j]=\left[\mathcal{U}^{D}[j], \mathcal{U}^{S}[j]\right]^T
\end{equation}
where
\begin{align}
\label{TCME_adj_gradient_matrice}
\frac{\partial\hspace{0.5mm}\mathcal{U}[j]}{\partial\hspace{0.5mm}\mathcal{U}[j-1]} 
&=
\begin{bmatrix}
    \frac{\partial\hspace{0.5mm}\mathcal{U}^{D}[j]}{\partial\hspace{0.5mm}\mathcal{U}^{D}[j-1]} & \frac{\partial\hspace{0.5mm}\mathcal{U}^{D}[j]}{\partial\hspace{0.5mm}\mathcal{U}^{S}[j-1]} \\
    \\[-1ex]
    \frac{\partial\hspace{0.5mm}\mathcal{U}^{S}[j]}{\partial\hspace{0.5mm}\mathcal{U}^{D}[j-1]} & \frac{\partial\hspace{0.5mm}\mathcal{U}^{S}[j]}{\partial\hspace{0.5mm}\mathcal{U}^{S}[j-1]}
\end{bmatrix} \notag \\
&=
\begin{bmatrix}
    \beta_1\beta_2+1 & \beta_1 \\
    \\[-1ex]
    \beta_1\beta_2^2+2\beta_2 & \beta_1\beta_2+1
\end{bmatrix}
\end{align}

In order to quantify the severity of the vanishing gradient problem in TC-LIF, we further calculate the 
column infinite norm as provided in Eq. (\ref{TCME_adj_gradient_inf_norm}) below. 

\begin{align}
\label{TCME_adj_gradient_inf_norm}
\norm{\frac{\partial\mathcal{U}[j]}{\partial\mathcal{U}[j-1]}}_{\infty}
&=\max(\beta_1\beta_2^2+\beta_1\beta_2+2\beta_2+1, \notag \\
&\phantom{=\max(}\beta_1\beta_2+\beta_1+1)  \\
&=\beta_1\beta_2^2+\beta_1\beta_2+2\beta_2+1 \notag
\end{align}
The infinite norm signifies the maximum changing rate of membrane potentials over a prolonged time period. By employing the constrained optimization method to solve the lower bound of the infinite norm for $\norm{\partial\hspace{0.5mm}\mathcal{U}[j]/\partial\hspace{0.5mm}\mathcal{U}[j-1]}_{\infty}$, it can be found that this value is larger than 1. This suggests that the TC-LIF model can effectively address the vanishing gradient problem. However, given that the value of infinite norm consistently exceeds 1, it inevitably faces the gradient exploding problem. Fortunately, our experimental studies show that this value is marginally greater than 1 for the majority of $\{\beta_1 , \beta_2\}$ selected from the second quadrant, thereby leading to stable training. See more analysis in Supplementary Materials.

It is worth noting that the TC-LIF model can be reformulated into a single-compartment form:
\begin{align}
\label{TCME_single_form}
\mathcal{U}^{S}[t] = &(1+\beta_1\hspace{0.5mm}\beta_2)\mathcal{U}^{S}[t-1] + \beta_2\hspace{0.5mm}\mathcal{U}^{D}[t-1] + \notag \\ &\beta_2\hspace{0.5mm}\mathcal{I}[t] - (\beta_2\gamma+\mathcal{V}_{th})\mathcal{S}[t-1] 
\end{align}

In essence, the above formulation mirrors a LIF neuron that is characterized by a decaying input. As a result, the proposed model is appropriately referred to as TC-LIF. Although the memory decaying problem persists in the TC-LIF model, the presence of $\mathcal{U}^D$ can effectively compensate for the memory loss and address the vanishing gradient problem.

\section{Experiments}
\label{sec: exp}
In this section, 
{we first explore the parameter space for the generalized two-compartment neurons to validate the design of TC-LIF.}
Then, we evaluate the TC-LIF model on various temporal classification benchmarks, including sequential MNIST (S-MNIST), permuted sequential MNIST (PS-MNIST), Google Speech Commands (GSC), Spiking Heidelberg Digits (SHD), and Spiking Google Speech Commands (SSC). 
Furthermore, we conduct a comprehensive study to demonstrate the superiority of the TC-LIF model in terms of remarkable temporal classification capability, effective long-term TCA, rapid training convergence, and high energy efficiency. Note that the training of TC-LIF model adopts the BPTT with surrogate gradients \cite{neftci2019surrogate}. More details about our experimental setups and training details are provided in Supplementary Materials.

\subsection{Parameter Space Exploration for Generalized Two-Compartment Neurons}
\label{sec: beta_study}
Based on the P-R model, we have put forward a generalized two-compartment neuron model whose neuronal dynamics are governed by $\{\alpha_1, \alpha_2, \beta_1, \beta_2\}$. As discussed earlier, to alleviate the rapid decay of memory stored in membrane potentials, we set both $\alpha_1$ and $\alpha_2$ to one in the TC-LIF model. It is worth noting that the initialization of $\beta_1$ and $\beta_2$ will, however, significantly affect the training convergence of a two-compartment neuron model. To shed light on the effectiveness of the proposed parameter setting in the TC-LIF model, we conduct a grid search by initializing $\beta_1$ and $\beta_2$ across four different quadrants and evaluate their performance on the S-MNIST and PS-MNIST datasets.

\begin{figure*}
    \centering
    \includegraphics[width=100mm]{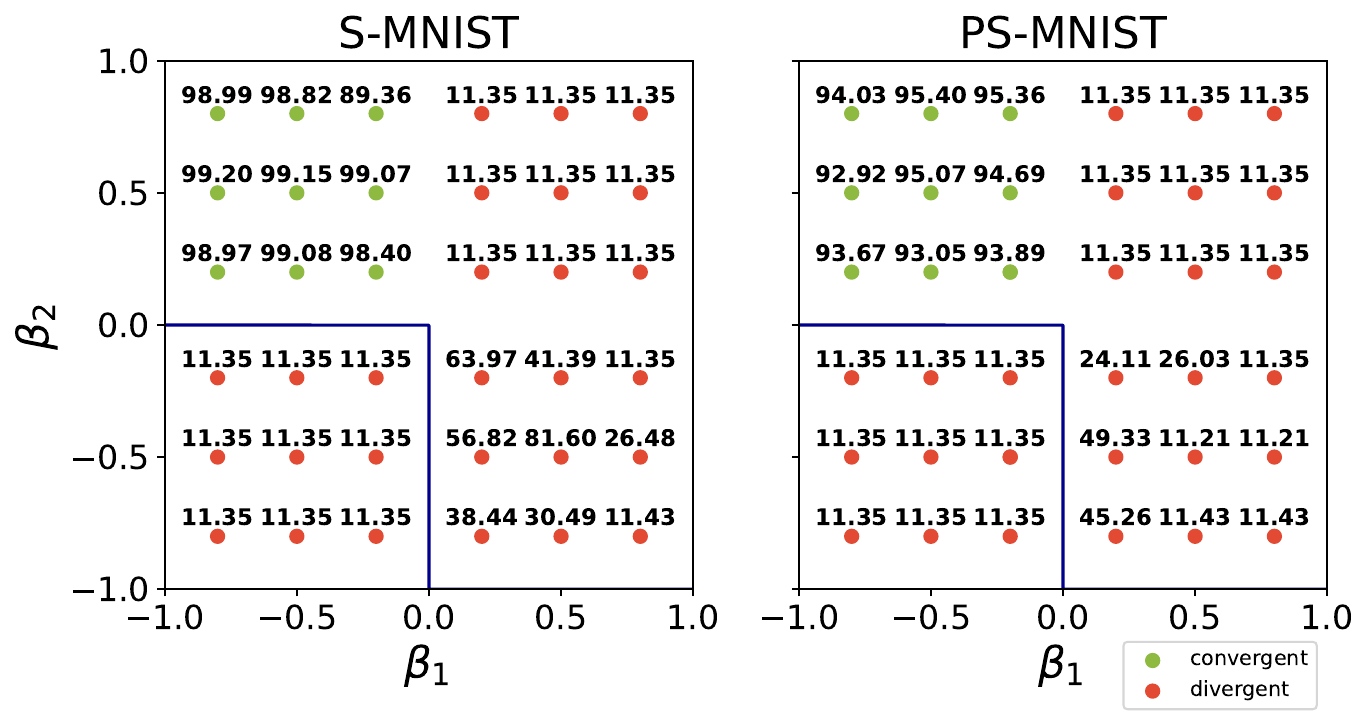}
    \caption{Study the impact of  $\beta_1$ and $\beta_2$ initialization on the test accuracy of S-MNIST and PS-MNIST datasets. Note that on both S-MNIST and PS-MNIST datasets, the models initialized in the first and third quadrants face severe exploding and vanishing gradient problems, respectively. As a result, they are unable to learn any meaningful information and are thus stuck at an accuracy of 11.35\%. The green dots refer to models that can converge to 100\% training accuracy.}
    \label{fig: beta_study}
\end{figure*}

\begin{figure*}[htb]
\centering
\subfloat[]{\includegraphics[width=55mm]{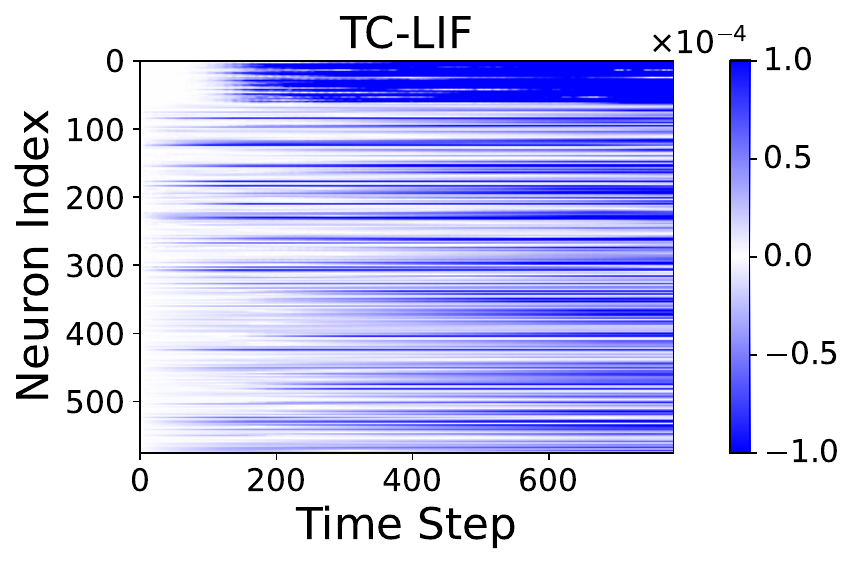}}
\quad
\subfloat[]{\includegraphics[width=55mm]{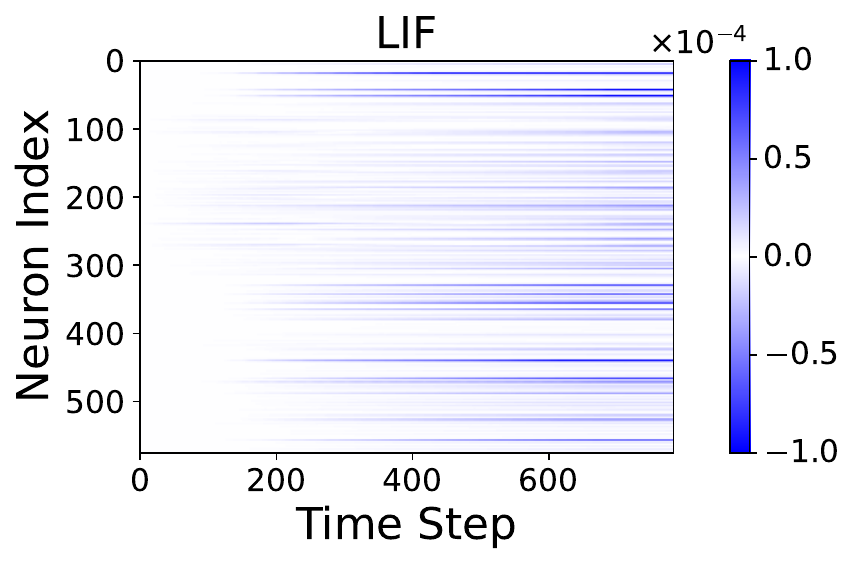}}
\quad
\subfloat[]{\includegraphics[width=55mm]{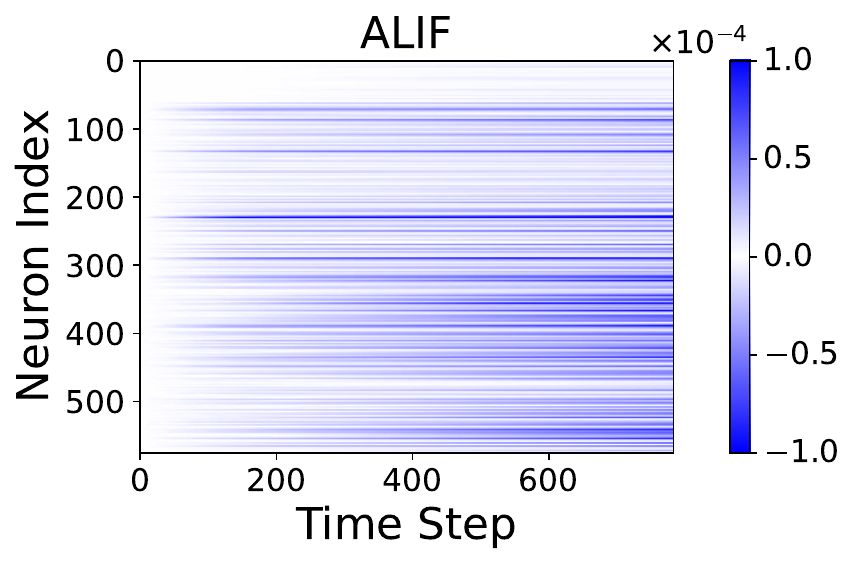}}
\caption{Illustration of the gradient evolution across time on the S-MNIST dataset. Note that the gradient has been calculated on a random batch of $256$ samples with a three-layer feedforward SNN (64-256-256).}
\label{fig: grads_prop}
\end{figure*}

In Figure \ref{fig: beta_study}, the dark blue line indicates the locations, where the partial derivative of the membrane potential between adjacent time steps $\frac{\partial\hspace{0.5mm}\mathcal{U}[j]}{\partial\hspace{0.5mm}\mathcal{U}[j-1]}$ equals to one. As a result, the entire space of $\beta$ is partitioned into two regions, wherein the third quadrant represents the region where the partial derivative is less than one. Conversely, the remaining three quadrants collectively represent the region where the partial derivative exceeds one. Within each quadrant, nine models are evaluated with equally spaced values for $\beta_1$ and $\beta_2$, and the numerical annotations adjacent to these points indicate the model's test accuracy on the respective dataset.

The result reveals that apart from the models initialized in the second quadrant, models in other regions struggle to converge. Particularly, when we initialize $\beta$ in the third quadrant, it refers to the scenario where the partial derivatives are less than 1 which will lead to the gradient vanishing problem. In contrast, models with $\beta$ initialized in the first quadrant face the severe gradient exploding problem. Both the gradient vanishing and exploding problems impede network convergence. 
Although initializing $\beta$ in the fourth quadrant can alleviate these problems, it results in a consistent negative input (see Eq. (\ref{TCME_soma})) to the somatic compartment that will lead to poor temporal classification results as seen across these two tasks. Therefore, we initialize the values of $\beta$ from the second quadrant, that is, $\beta_1 \in (-1, 0)$ and $\beta_2 \in (0, 1)$ for our TC-LIF model and we use it consistently for the rest of our experiments.


\subsection{Superior Temporal Classification Capability}
\label{sec: comp_sota}
Table \ref{tab: comp_sota} presents the results of the proposed TC-LIF model on five commonly used temporal classification datasets, along with other existing works. Overall, the TC-LIF model consistently outperforms SOTA single-compartment neuron models with a comparable amount of parameters.

\begin{table*}[ht!]
  \centering
  \small
  \resizebox{0.9\textwidth}{!}{%
  \begin{tabular}{clcccc}
    \toprule
    \textbf{Datasets}   &\textbf{Method}  &\textbf{SNN}   &\textbf{Network}   &\textbf{Parameters (K)}    &\textbf{Accuracy (\%)} \\
    \midrule
    \multirow{16}{*}{\rotatebox{90}{S-MNIST}}      & GLIF* \cite{yao2022glif}  & Y   & FF       & 47.1/87.5  & 94.80/95.27 \\               & PLIF* \cite{fang2021incorporating}  & Y   & FF     & 44.8/85.1  & 83.71/87.92 \\
                  & LIF*  & Y    & FF         & 44.8/85.1         &62.42/72.06 \\
                  & LTMD* \cite{wang2022ltmd}    & Y            & FF       & - /85.1  & - /68.56 \\
                  & \textbf{TC-LIF (ours)}  & Y  & \textbf{FF} & \textbf{44.8/85.1}   & \textbf{96.46/97.35} \\
    \cmidrule(r){2-6}
                  & LSTM  \cite{arjovsky2016unitary} & N  & Rec         & 66.5/ -        & 98.20/ - \\
                  & SRNN+ReLU  \cite{yin2020effective}    & N     & Rec         & 129.6/ -         & 98.99/ - \\
                  & LSNN \cite{bellec2018long}       & Y       & Rec         & 68.2/ -        & 93.70/ - \\
                  & AHP \cite{rao2022long}       & Y       & Rec         & 68.4/ -        & 96.00/ - \\
                  & GLIF* \cite{yao2022glif}      & Y      & Rec       & 114.6/157.5     & 95.63/96.64 \\
                  & SRNN+ALIF \cite{yin2020effective,yin2021accurate}  & Y  & Rec         & 129.6/156.3         & 97.82/98.70 \\
                  & PLIF* \cite{fang2021incorporating}  & Y  & Rec     & 112.2/155.1  & 90.93/91.79 \\
                  & LIF*               & Y             & Rec         & 112.2/155.1         & 74.91/89.28 \\
                  & LTMD* \cite{wang2022ltmd}   & Y            & Rec       & - /155.1  & - /84.62 \\
                  & \textbf{TC-LIF (ours)} & Y  &  \textbf{Rec}  &  \textbf{63.6/155.1}  &  \textbf{98.79/99.20} \\
    \bottomrule
    \multirow{10}{*}{\rotatebox{90}{PS-MNIST}}   
                  & LIF*  & Y    & FF         & 44.8/85.1         & 11.30/10.00 \\
                  & \textbf{TC-LIF (ours)}  & Y &  \textbf{FF} & \textbf{44.8/85.1}   & \textbf{80.89/83.98} \\
    \cmidrule(r){2-6}
                  & LSTM  \cite{arjovsky2016unitary}   & N   & Rec         & 66.5/ -       & 88.00/ - \\
                  & SRNN+ReLU  \cite{yin2020effective}   & N      & Rec         & 129.6/ -         & 93.47/ - \\
                  & GLIF* \cite{yao2022glif}      & Y       & Rec         & 114.6/157.5         & 90.34/90.47 \\
                  & SRNN+ALIF \cite{yin2020effective,yin2021accurate}   & Y      & Rec         & 129.6/156.3         & 91.00/94.30 \\
                  & LIF*  & Y    & Rec         & 112.2/155.1         & 71.77/80.39 \\
                  & LTMD* \cite{wang2022ltmd}    & Y            & Rec       & - /155.1  & - /54.93 \\                            
                  & \textbf{TC-LIF (ours)} & Y &  \textbf{Rec}  &  \textbf{63.6/155.1} &  \textbf{92.69/95.36} \\
    \bottomrule
    \multirow{7}{*}{\rotatebox{90}{GSC}}   & Rate-based SNN \cite{yilmaz2020deep}  & Y  & FF     & 117       & 75.20 \\
                  & \textbf{TC-LIF (ours)} & Y & \textbf{FF} & \textbf{106.2}   & \textbf{91.35} \\
    \cmidrule(r){2-6} 
                  & SRNN+ALIF \cite{yin2021accurate}   & Y      & Rec         & 221.7         & 92.10 \\
                  & SNN \cite{salaj2021spike}       & Y       & Rec         & 4304.9        & 89.04 \\
                  & SNN with SFA \cite{salaj2021spike}  & Y    & Rec         & 4307          & 91.21 \\
                  & \textbf{TC-LIF (ours)} & Y &  \textbf{Rec}  &  \textbf{196.5} &  \textbf{94.84} \\
    \bottomrule
    \multirow{10}{*}{\rotatebox{90}{SHD}}   & Feed-forward SNN \cite{cramer2020heidelberg}  & Y  & FF     & 108.8       & 48.60 \\
                  & \textbf{TC-LIF (ours)} & Y & \textbf{FF} & \textbf{108.8}   & \textbf{83.08} \\
    \cmidrule(r){2-6}
                  & SRNN \cite{cramer2020heidelberg}      & Y         & Rec         & 108.8        & 71.4 \\
                  & Heterogeneous SRNN \cite{perez2021neural} & Y & Rec       & 108.8      & 82.70 \\
                  & Attention \cite{yao2021temporal}   & Y       & Rec         & 133.8        & 81.45 \\
                  & SRNN + ALIF \cite{yin2020effective}   & Y     & Rec         & 142.4        & 84.40 \\
                  & SRNN  \cite{zenke2021remarkable}       & Y        & Rec         & 249        & 82.00 \\
                  & SRNN + data augm. \cite{cramer2020heidelberg}  & Y  & Rec      & 1787.9       & 83.20 \\
                  & \textbf{TC-LIF (ours)}  & Y   & \textbf{Rec}  &  \textbf{141.8} &  \textbf{88.91} \\
    \bottomrule
    \multirow{6}{*}{\rotatebox{90}{SSC}}   & Feed-forward SNN \cite{cramer2020heidelberg}  & Y  & FF     & 110.8       & 38.50 \\
                  & \textbf{TC-LIF (ours)}  & Y  & \textbf{FF} & \textbf{110.8}   & \textbf{63.46} \\
    \cmidrule(r){2-6}
                  & SRNN \cite{cramer2020heidelberg}        & Y         & Rec       & 110.8      & 50.90 \\
                  & Heterogeneous SRNN \cite{perez2021neural}   & Y   & Rec       & 110.8      & 57.3 \\
                  & \textbf{TC-LIF (ours)}  & Y  &  \textbf{Rec}  &  \textbf{110.8} &  \textbf{61.09} \\
    \bottomrule
  \end{tabular}}
  \caption{Comparison of model performance on S-MNIST, PS-MNIST, GSC, SHD, and SSC datasets. Here, "FF" and "Rec" represent feedforward and recurrent networks, respectively. * denotes our reproduced results using publicly available codes.}
  \label{tab: comp_sota}
\end{table*}

For the S-MNIST dataset, each data sample has a sequence length of 784, which requires the model to learn long-term dependencies. As expected, the LIF model performs worst on this dataset, which can be explained by the vanishing gradient problem discussed earlier. Notably, the recently introduced adaptive spiking neuron model: LSNN \cite{bellec2018long} and adaptive LIF (ALIF) \cite{yin2020effective, yin2021accurate} achieve comparable or even better accuracies to the LSTM model \cite{arjovsky2016unitary}. Our proposed TC-LIF model consistently outperforms these single-compartment neuron models, indicating its effectiveness in retaining multiscale temporal information and handling long-term dependencies. Notably, we achieve 
99.20\% accuracy with a recurrent architecture. To the best of our knowledge, this is the best-reported SNN model on this dataset. The same conclusions can also be drawn for the more challenging PS-MNIST dataset.

In addition to image datasets, we further conduct experiments on speech datasets that exhibit richer temporal dynamics. For the non-spiking GSC dataset, our TC-LIF model achieves 91.35\% and 94.84\% accuracy for feedforward and recurrent networks respectively, surpassing SOTA models by a large margin. The SHD and SSC datasets are neuromorphic datasets that are specifically designed for benchmarking SNNs. On these datasets, our proposed TC-LIF exhibit a significant improvement over all other reported works.

\subsection{Effective Long-term Temporal Credit Assignment}
To provide more insights into how long-term temporal relationship has been established in TC-LIF neurons, we visualize the error gradients calculated on the S-MNIST dataset. To enhance visual clarity, the gradient value $\mathcal{G}^{n}_{t}$ of neuron $n$ at time step $t$ is normalized as $\mathcal{G}^{n}_{t}/\sum^{N}_{i=0}\sum^{T}_{j=0}\mathcal{G}^{i}_{j}$. 

\begin{figure*}[htb]
\centering
\subfloat[]{\includegraphics[width=60mm]{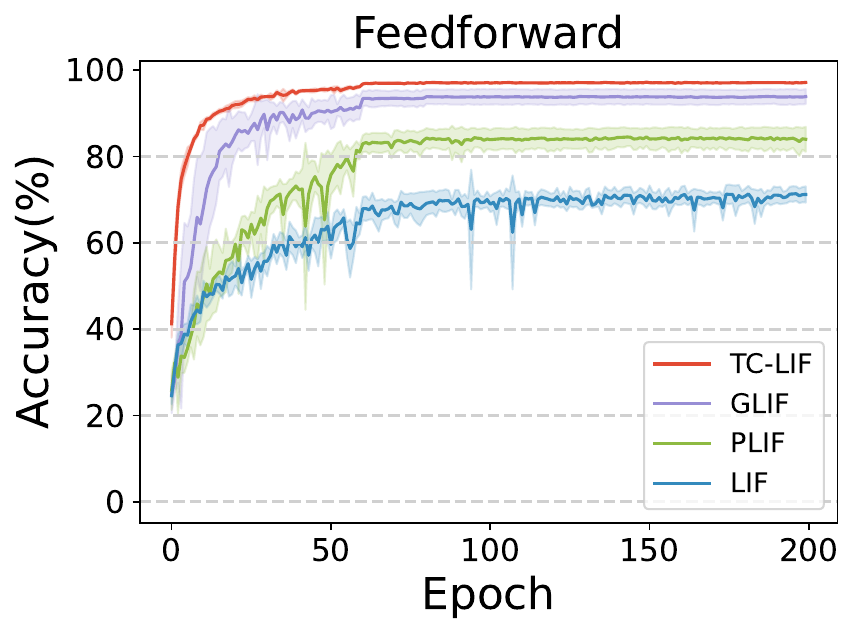}} \quad
\subfloat[]{\includegraphics[width=60mm]{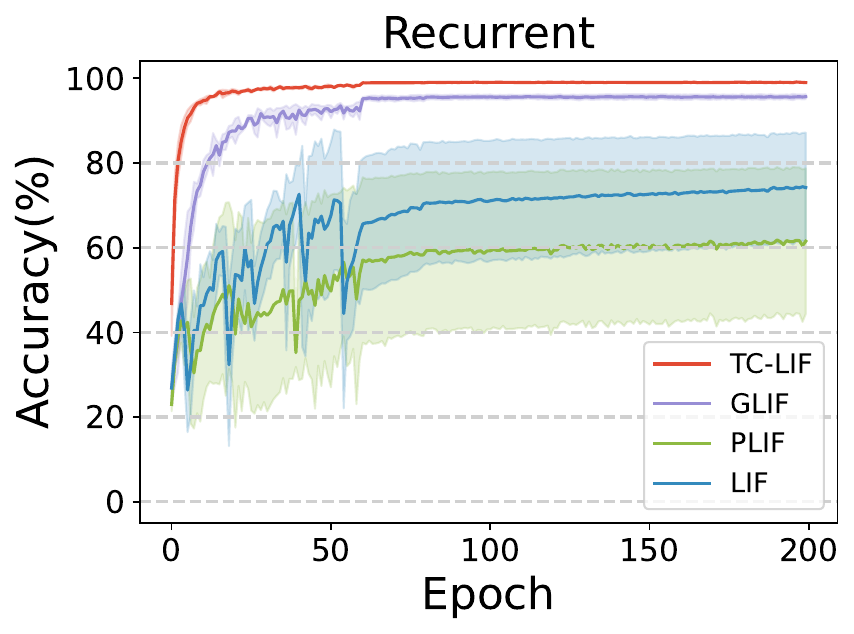}} 
\caption{Comparison of the learning curves of TC-LIF and other single-compartment spiking neurons with (a) feedforward and (b) recurrent network architectures. Note that the mean and standard deviations across four runs are reported.}
\label{fig: lr_curve}
\end{figure*}

\begin{table*}[ht!]
  \centering
  \resizebox{0.8\textwidth}{!}{%
  \begin{tabular}{lcc}
    \toprule
    \textbf{ Neuron Model}   &\textbf{ Theoretical Energy Cost}  &\textbf{ Empirical Energy Cost (nJ)} \\
    \midrule
    
    \multirow{2}{*}{LSTM}            & { $4(mn+nn)E_{MAC}$}                       &\multirow{2}{*}{ 2,834.7}\\
                    & { $+17nE_{MAC}$}                          &           \\
    
    \multirow{2}{*}{LIF}     & { $mnFr_{in}E_{AC}+(nn+n)Fr_{out}E_{AC}$}                             & \multirow{2}{*}{ 23.8}\\
            & { $+nE_{MAC}$}              &\\
    
    \multirow{2}{*}{TC-LIF}          & { $mnFr_{in}E_{AC}+(nn+2n)Fr_{out}E_{AC}$}  & \multirow{2}{*}{ 28.2}\\
                  & {$+2nE_{MAC}$}   & \\
    \bottomrule
  \end{tabular}}
  \caption{Comparison of the theoretical and empirical energy cost of LIF, LSTM, and TC-LIF models. The $m$ and $n$ are the numbers of input and output neurons. $Fr_{in}$ and $Fr_{out}$ are the firing rate of input and output neurons. $E_{AC}$ and $E_{MAC}$ are the energy cost of MAC and AC operations.}
  \label{tab: energy_comp}
\end{table*}

\begin{figure*}[htb]
\centering
\subfloat[]{\includegraphics[width=60mm]{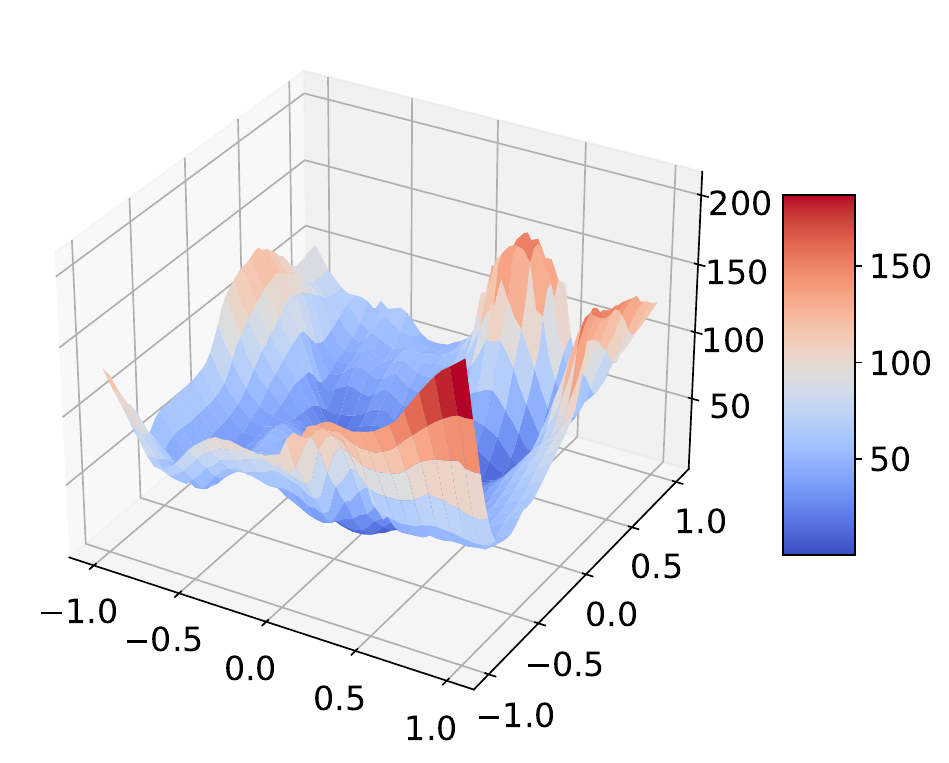}} 
\subfloat[]{\includegraphics[width=60mm]{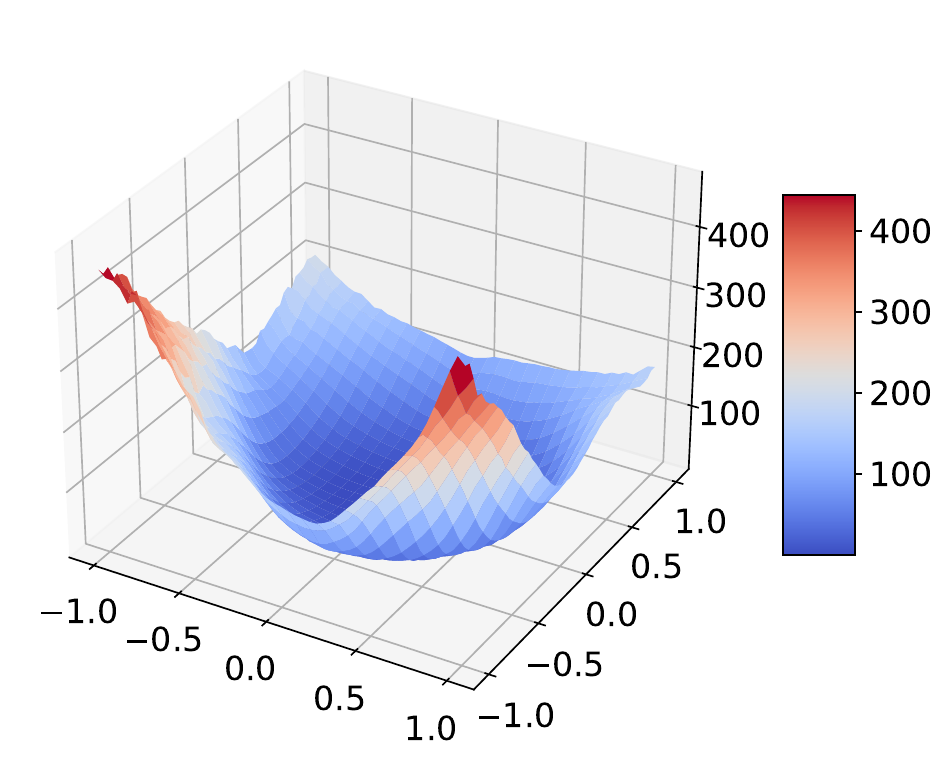}} 
\\
\subfloat[]{\includegraphics[width=57mm]{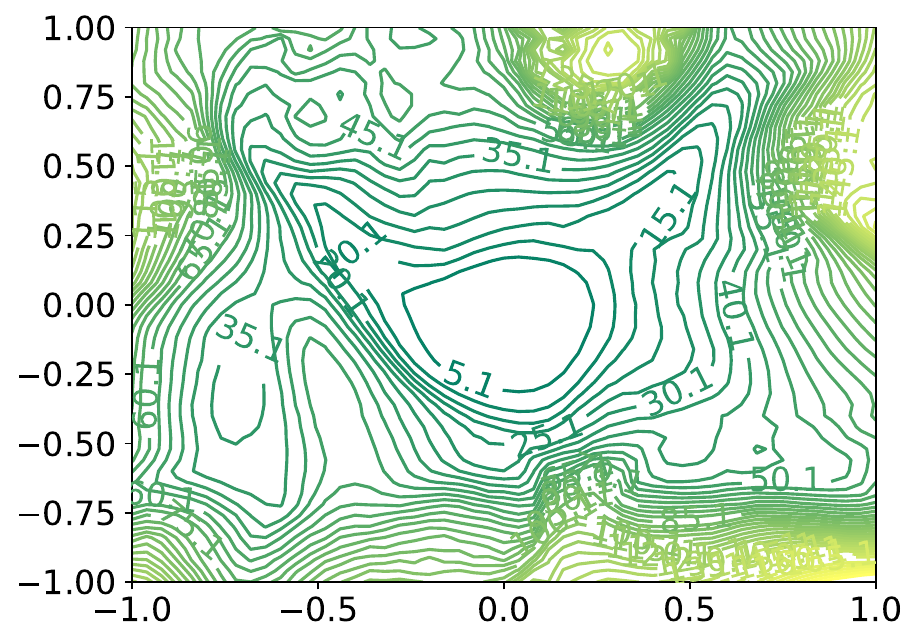}} 
\subfloat[]{\includegraphics[width=57mm]{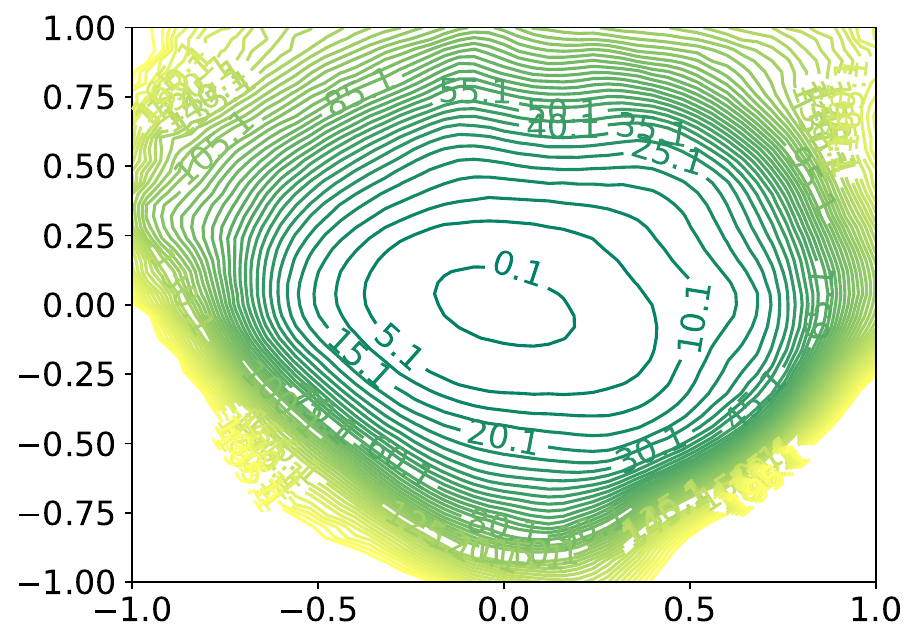}} 
\caption{Comparison of the loss landscape of (a, c) LIF and (b, d) TC-LIF neuron models in terms of 3D surface and 2D contour plots.}
\label{fig: loss_landscape}
\end{figure*}

As presented in Figure \ref{fig: grads_prop}, TC-LIF neurons can effectively deliver more gradients to the earlier time steps as compared with LIF and ALIF models. This is more evident for the first and second layers (Neuron Index $0-319$). These results suggest the exceptional ability of TC-LIF in performing long-term TCA.

\subsection{Rapid Training Convergence}
\label{sec: convergence}
Taking benefits of the exception ability in performing long-term TCA, the proposed TC-LIF model ensures more stable learning and faster network convergence. To shed light on this, we further compare the learning curves of the TC-LIF model alongside the LIF, GLIF, and PLIF models under the same training condition.
As illustrated in Figure \ref{fig: lr_curve}, the solid line denotes the mean accuracy, while the shaded area encapsulates the accuracy standard deviation across four runs with different random seeds. Notably, the TC-LIF model converges rapidly within about 25 epochs for both network structures, while the LIF model takes around 100 and 75 epochs to converge for feedforward and recurrent networks, respectively. 
Moreover, the TC-LIF model exhibits greater stability than other models, especially during the early training stage. 

Furthermore, to investigate the reason why the TC-LIF model can achieve more stable learning and faster convergence, we further compare the loss landscape of LIF and TC-LIF near the found local minima.
As shown in Figure \ref{fig: loss_landscape}, it is obvious that the TC-LIF model exhibits a notably smoother loss landscape near the local minima. This suggests the TC-LIF model offers improved learning dynamics and convergence properties.
In particular, the smoother loss landscape enables a reduced likelihood of being trapped into local minima, which can lead to more stable optimization and faster convergence. Additionally, the smoother loss landscape suggests stronger network generalization, as it is less prone to overfitting and underfitting problems. 

\subsection{High Energy Efficiency}
\label{sec: efficient}
So far, it remains unclear whether the proposed TC-LIF model can make a good trade-off between model complexity and computational efficiency. To answer this question, we analyze and compare the energy efficiency of LIF, TC-LIF, and LSTM models. In particular, we count the accumulated (AC) and multiply-and-accumulate (MAC) operations consumed during input data processing and network update. In ANNs, the computations are all performed with MAC operations, whereas the AC operations are used predominantly in SNNs for synaptic updates. It is worth noting that the membrane potential update of spiking neurons requires several MAC operations. More detailed calculations can be found in Supplementary Materials.

As the theoretical results presented in Table \ref{tab: energy_comp}, the energy costs of both spiking neurons (i.e., LIF and TC-LIF) are significantly lower than that of the LSTM model, attributed to their reduced computational complexity. Compared to the LIF model, the proposed TC-LIF model incurs additional $nFr_{out}E_{AC}+nE_{MAC}$ operations due to the extra computations at the dendritic compartment. 
To calculate the empirical energy cost, we perform inference on one randomly selected batch of test samples and compute the average layer-wise firing rates of these SNNs on the S-MNIST dataset. The layer-wise firing rates for LIF and TC-LIF models are comparable that take the values of [0.219, 0.145, 0.004] and [0.294, 0.146, 0.030], respectively. To obtain the total energy cost, we base our calculation on the 45nm CMOS process that has an estimated cost of $E_{AC}=0.9~pJ$ and $E_{MAC}=4.6~pJ$ for AC and MAC operations, respectively \cite{horowitz20141}. Despite the more complex internal structure of the proposed TC-LIF model, it has a comparable energy cost to the LIF model. Remarkably, our TC-LIF model achieves more than 100 times energy savings compared with the LSTM model, while demonstrating superior temporal classification performance.

\section{Conclusion}
\label{sec: conclusion}
In this paper, drawing inspiration from the multi-compartment structure of biological neurons, we proposed a novel TC-LIF neuron to enhance the long-term sequential modeling capability of spiking neurons. The dendritic and somatic compartments of the proposed TC-LIF model synergistically interact, enriching neuronal dynamics and effectively addressing the TCA problem when properly configured. Theoretical analysis and experimental results on various temporal classification tasks demonstrate the superiority of the proposed TC-LIF model, including superior temporal classification capability, effective long-term TCA, rapid training convergence, and high energy efficiency. This work, therefore, contributes to the development of more effective and efficient spiking neurons for solving sequential modeling tasks. In this work, we focus our study on two-compartment neuron models, while how to generalize the design to more compartments remains an interesting question that we will explore in future works.

\section{Acknowledgments}
This work was supported in part by the Research Grants Council of the Hong Kong SAR (Grant No. PolyU11211521, PolyU15218622, and PolyU25216423), The Hong Kong Polytechnic University (Project IDs: P0039734, P0035379, P0043563, and P0046094), the National Natural Science Foundation of China (Grant No. U21A20512, 62306259, 62271432), the IAF, A*STAR, SOITEC, NXP and National University of Singapore under FD-fAbrICS: Joint Lab for FD-SOI Always-on Intelligent Connected Systems (Award I2001E0053), the Agency for Science, Technology and Research (A*STAR) under its AME Programmatic Funding Scheme (Project No. A18A2b0046), and A*STAR under its RIE 2020 Advanced Manufacturing and Engineering Human (AME) Programmatic Grant (Grant No. A1687b0033).

\bibliography{aaai24}

\clearpage

\section{Supplementary Materials}
\appendix
\section{Pinsky-Rinzel Neuron Model}
\label{app: method}

This section presents the generalized two-compartment spiking neuron model derived from the two-compartment Pinsky-Rinzel (P-R) pyramidal neuron model. The two-compartment P-R neuron model is designed to elucidate intricate biophysical mechanisms that underly complex bursting within CA3 pyramidal cells and enables lightweight computation. Its neuronal dynamic can be formulated in continuous time by the following equations:

\begin{equation}
    \label{eq-supp: continuous_soma}
    C_{m}\frac{dV_{s}}{dt}=-I_{Na}-I_{K}-I_{Leak}+\frac{I_{link}}{P}+I_{s}
\end{equation}

\begin{equation}
    \label{eq-supp: continuous_dendrite}
    C_{m}\frac{dV_{d}}{dt}=-I_{NaP}-I_{KS}-I_{Leak}-\frac{I_{link}}{1-P}+I_{d}
\end{equation}

where $V_s$ and $V_d$ are the membrane potentials of somatic and dendritic compartments. $I_{Na}$ and $I_{K}$ denote the pertinent currents in the somatic compartment, while the dendritic compartment encompasses the slow potassium current $I_{KS}$ and persistent sodium current $I_{NaP}$. The input currents to the soma and dendrite are denoted by $I_s$ and $I_d$ respectively. Specifically, $I_s$ is assumed to be 0 in this paper and the input currents are solely injected into the dendritic compartment. Additionally, the membrane capacitance and the proportion of cell area are represented by $C_m$ and $P$ respectively.

Table \ref{tab: P-R_ionic_currents} provides the detailed calculations related to the ionic currents mentioned in the equations above. In particular, $E_{Na}$, $E_{K}$, and $E_{L}$ signify equilibrium potentials, while $g_{Na}$, $g_{K}$, $g_{L}$, $g_{c}$, $g_{NaP}$, and $g_{KS}$ represent conductances.

\begin{table}[htb]
  \caption{Summary of the ionic current calculation in two-compartment P-R neuron model.}
  \label{tab: P-R_ionic_currents}
  \centering
  \begin{tabular}{cc}
    \toprule
    \textbf{Ionic Current} &\textbf{Calculation}\\
    \midrule
    $I_{Na}$    & $g_{Na}m^3h\cdot (V_s-E_{Na})$ \\
    $I_{K}$     & $g_{K}n^4\cdot (V_s-E_{K})$     \\
    $I_{Leak}$  & $g_{L}\cdot (V-E_{L})$    \\
    $I_{link}$ & $g_{c}\cdot (V_{d}-V_{s})$ \\ 
    $I_{NaP}$  & $g_{NaP}l^3h\cdot (V_d-E_{Na})$ \\
    $I_{KS}$   & $g_{KS}q\cdot (V_d-E_{K})$ \\
    \bottomrule
  \end{tabular}
\end{table}

We obtain the discrete-time formulations for Eqs. (\ref{eq-supp: continuous_soma}) and (\ref{eq-supp: continuous_dendrite}) by the Euler method as follows:

\begin{align}
    \label{eq-supp: discrete_soma}
    V_{s}[t+1] = &V_{s}[t]+ \frac{dt}{C_m}(-I_{Na}[t]-I_{K}[t]- \notag \\
    &I_{Leak}[t]+\frac{I_{link}[t]}{P})
\end{align}

\begin{align}
    \label{eq-supp: discrete_dendrite}
    V_{d}[t+1] = &V_{d}[t]+\frac{dt}{C_m}(-I_{NaP}[t]-I_{KS}[t]- \notag
    \\
    &I_{Leak}[t]+\frac{I_{link}[t]}{1-P}+I_{d}[t])
\end{align}

The term $I_{link}$ signifies the interaction between the somatic and dendritic compartments. Additionally, by substituting the expressions for ionic currents from the Table \ref{tab: P-R_ionic_currents} into Eq. (\ref{eq-supp: discrete_soma}) and (\ref{eq-supp: discrete_dendrite}) and integrating the spiking operation for the somatic output membrane potential, we derive the overall dynamics of the generalized two-compartment spiking neuron model, as depicted in Eq. (\ref{PR_dendrite_general})-(\ref{PR_spike_cal}).

\section{Energy Efficiency Analysis}
\label{app: energy_study}
We analyze the theoretical energy cost for LSTM, LIF, and TC-LIF recurrent networks based on their neuronal update functions.
Table \ref{tab: energy cost} presents the detailed calculation of theoretical energy cost for each model. 


\begin{table*}[htb]
  \caption{Computations on the energy cost of LIF, TC-LIF, and LSTM.}
  \label{tab: energy cost}
  \centering
  \resizebox{0.8\textwidth}{!}{%
  \begin{tabular}{lccc}
    \toprule
    \textbf{Neuron Model}  &\textbf{Dynamics} & \textbf{Step Cost}  & \textbf{Total Cost}   \\
    \midrule
    \multirow{2}{*}{LIF}            & $\mathcal{I}_{t}=\mathcal{W}^{m,n}X^m+\mathcal{W}^{n,n}\mathcal{S}^n_{t-1}$     &      $(mnFr_{in}+nnFr_{out})E_{AC}$      & $mnFr_{in}E_{AC}+(nn+n)Fr_{out}E_{AC}$    \\
                                    & $\mathcal{U}_{t}=\beta\mathcal{U}_{t-1}+\mathcal{I}_{t}-\mathcal{V}_{th}\mathcal{S}^n_{t-1}$     & $nFr_{out}E_{AC}+nE_{MAC}$      & $+nE_{MAC}$    \\
    \midrule
    \multirow{3}{*}{TC-LIF}       & $\mathcal{I}_{t}=\mathcal{W}^{m,n}X^m+\mathcal{W}^{n,n}\mathcal{S}^n_{t-1}$ & $(mnFr_{in}+nnFr_{out})E_{AC}$      & $mnFr_{in}E_{AC}$       \\
                                    & $\mathcal{U}^{D}_{t}=\mathcal{U}^{D}_{t-1}+\mathcal{I}_{t}+\beta_1\mathcal{U}^{S}_{t-1}-\gamma\mathcal{S}^n_{t-1}$  & $nFr_{out}E_{AC}+nE_{MAC}$      & $+(nn+2n)Fr_{out}E_{AC}$       \\
                                    & $\mathcal{U}^{S}_{t}=\mathcal{U}^{S}_{t-1}+\beta_2\mathcal{U}^{D}_{t}-\mathcal{V}_{th}\mathcal{S}^n_{t-1}$& $nFr_{out}E_{AC}+nE_{MAC}$&$+2nE_{MAC}$ \\
    \midrule
    \multirow{6}{*}{LSTM}           & $f_{t}=\sigma_g(\mathcal{W}_{f}x_{t}+{U}_{f}h_{t-1}+b_{f})$        & $n(m+n+2)E_{MAC}$        &                      \\
                                    & $i_{t}=\sigma_g(\mathcal{W}_{i}x_{t}+{U}_{i}h_{t-1}+b_{i})$        & $n(m+n+2)E_{MAC}$        &                      \\
                                    & $o_{t}=\sigma_g(\mathcal{W}_{o}x_{t}+{U}_{o}h_{t-1}+b_{o})$        & $n(m+n+2)E_{MAC}$        & $4(mn+nn)E_{MAC}$    \\
                                    & $\hat{c}_{t}=\sigma_c(\mathcal{W}_{c}x_{t}+{U}_{c}h_{t-1}+b_{c})$  & $n(m+n+4)E_{MAC}$        & $17nE_{MAC}$         \\
                                    & $c_{t}=f_{t} \odot c_{t-1}+i_{t} \odot \hat{c}_{t}$                & $2nE_{MAC}$              &                      \\
                                    & $h_{t}=o_{t} \odot \sigma_{h}(c_{t})$                              & $5nE_{MAC}$              &                      \\

    \bottomrule
  \end{tabular}}
\end{table*}

\section{Experimental Details}
\label{app: exp_setting}

\subsection{Datasets}
\label{app: datasets}
In this subsection, we introduce the dataset used for this work. These datasets cover a wide range of tasks, allowing us to assess the model's capabilities in handling different types of input data.

\textbf{S-MNIST:} The Sequential-MNIST (S-MNIST) dataset is derived from the original MNIST dataset, which consists of 60,000 and 10,000 grayscale images of handwritten digits for training and testing sets with a resolution of 28 $\times$ 28 pixels. In the S-MNIST dataset, each image is converted into a vector of 784 time steps, with each pixel representing one input value at a certain time step. This dataset enables us to evaluate the performance of our model in solving sequential image classification tasks.

\textbf{PS-MNIST:} The Permuted Sequential MNIST dataset (PS-MNIST) is a variation of the Sequential MNIST dataset, in which the pixels in each image are shuffled according to a fixed random permutation. This dataset provides a more challenging task than S-MNIST, as the input sequences no longer follow the original spatial order of the images. Therefore, when learning this dataset, the model needs to capture complex, non-local, and long-term dependencies between pixels. 

\textbf{GSC:} The Google Speech Commands (GSC) has two versions, and we employ the 2nd version in this work. The GSC version 2 is a collection of 105,829 on-second-long audio clips of 35 different spoken commands, such as “yes”, “no”, “up”, “down”, “left”, “right”, etc. These audio clips are recorded by different speakers in various environments, offering a diversity of datasets to evaluate the performance of our model.

\textbf{SHD:} The Spiking Heidelberg Digits dataset is a spike-based sequence classification benchmark, consisting of spoken digits from 0 to 9 in both English and German (20 classes). The dataset contains recordings from twelve different speakers, with two of them only appearing in the test set. Each original waveform has been converted into spike trains over 700 input channels. The train set contains 8,332 examples, and the test set consists of 2,088 examples (no validation set). The SHD dataset enables us to evaluate the performance of our proposed model in processing and classifying speech data represented in spiking format.

\textbf{SSC:} The Spiking Speech Command dataset, another spike-based sequence classification benchmark, is derived from the Google Speech Commands version 2 dataset and contains 35 classes from a large number of speakers. The original waveforms have been converted to spike trains over 700 input channels. The dataset is divided into train, validation, and test splits, with 75,466, 9,981, and 20,382 examples, respectively. The SSC dataset allows us to assess the performance of our proposed spiking neuron model in processing and recognizing speech commands represented in spiking data.

\subsection{Training with Surrogate Gradient}
\label{app: surrogate}
Training SNN poses challenges stemming from the non-differentiability of spike functions, denoted as $\Theta(x)$ in Eqs. (\ref{Surrogate_spike_cal}), (\ref{PR_spike_cal}), and (\ref{TCME_spike_cal}). This trait hinders the application of prevalent gradient-based optimization methods, notably backpropagation. 
The surrogate gradient approach offers a solution to this impediment by introducing a proxy gradient as an approximation for the gradient of the spike function, expressed as $\Theta'(x) \approx \theta'(x)$. While the actual gradient mostly holds a zero value, the surrogate gradient approximates non-zero values in regions of interest. This allows backpropagation to be applied, as the surrogate gradient provides the necessary feedback to update the network's weight.

In this work, we adopt the triangle function as $\theta'(x)$ to enable gradient-based training for SNN:
\begin{equation}
    \label{eq-supp: surrogate}
    \frac{\partial \mathcal{S}[t]}{\partial \mathcal{U}[t]} = \theta'(\mathcal{U}[t] - \mathcal{V}_{th}) = \max(1- \left| \mathcal{U}[t] - \mathcal{V}_{th} \right|, 0)
\end{equation}
where $\mathcal{U}[t]$ denotes the membrane potential in Eq. (\ref{Surrogate_spike_cal}) for single-compartment spiking neuron and the somatic membrane potential in Eqs. (\ref{PR_spike_cal}) and (\ref{TCME_spike_cal}) for two-compartment spiking neurons.

\subsection{Network Architecture}
\label{app: configure}
We perform experiments employing both feedforward and recurrent connection configurations. To maintain a fair comparison with existing works, we utilize network architectures exhibiting comparable parameters. These architectures and their corresponding parameters are summarized in Table \ref{tab: arch}.

\subsection{TC-LIF Model Hyper-parameters}
\label{app: hp}
We outline the specific hyper-parameter settings for the TC-LIF neuron model in Table \ref{tab: hp}, encompassing the dendritic reset scalar $\gamma$, the spike threshold $\mathcal{V}_{th}$, and initial values for $\beta_1$ and $\beta_2$.

\begin{table}[htb]
  \caption{Network hyper-parameters for TC-LIF.}
  \label{tab: hp}
  \centering
  \resizebox{0.45\textwidth}{!}{%
  \begin{tabular}{lcccc}
    \toprule
    \textbf{Dataset}  &\textbf{Network} & \textbf{$\gamma$}  & $\beta_1$, $\beta_2$  & $\mathcal{V}_{th}$ \\
    \midrule
    \multirow{2}{*}{S-MNIST}        & feedforward   & 0.5      & (-0.5, 0.5)   & 1.0  \\
                                    & recurrent     & 0.5      & (-0.8, 0.4)   & 1.0 \\
    \midrule
    \multirow{2}{*}{PS-MNIST}       & feedforward   & 0.7      & (-0.5, 0.5)   & 1.5 \\
                                    & recurrent     & 0.5      & (-0.2, 0.8)   & 1.8 \\
    \midrule
    \multirow{2}{*}{GSC}            & feedforward   & 0.6       & (-0.5, 0.5)  & 1.2  \\
                                    & recurrent     & 0.7       & (-0.8, 0.8)  & 1.25  \\
    \midrule
    \multirow{2}{*}{SHD}            & feedforward   & 0.5    & (-0.5, 0.5)  & 1.5  \\
                                    & recurrent     & 0.5   & (-0.5, 0.5)  & 1.5  \\
    \midrule
    \multirow{2}{*}{SSC}            & feedforward   & 0.5    & (-0.5, 0.5)  & 1.5  \\
                                    & recurrent     & 0.5    & (-0.5, 0.5)  & 1.5  \\
    \bottomrule
  \end{tabular}}
\end{table}

\begin{table}[ht!]
  \caption{Summary of network architectures and parameters.}
  \label{tab: arch}
  \centering
  \resizebox{0.45\textwidth}{!}{%
  \begin{tabular}{lccc}
    \toprule
    \textbf{Dataset}  &\textbf{Network} & \textbf{Architecture}  & \textbf{Parameters(K)}   \\
    \midrule
    \multirow{2}{*}{S-MNIST}        & feedforward   & 40-256-128-10/64-256-256-10      & 44.8/85.1    \\
                                    & recurrent     & 40-200-64-10/64-256-256-10      & 63.6/155.1    \\
    \midrule
    \multirow{2}{*}{PS-MNIST}       & feedforward   & 40-256-128-10/64-256-256-10      & 44.8/85.1    \\
                                    & recurrent     & 40-200-64-10/64-256-256-10      & 63.6/155.1    \\
    \midrule
    \multirow{2}{*}{GSC}            & feedforward   & 40-300-300-12       & 106.2    \\
                                    & recurrent     & 40-300-300-12       & 196.5    \\
    \midrule
    \multirow{2}{*}{SHD}            & feedforward   & 700-128-128-20     & 108.8    \\
                                    & recurrent     & 700-128-128-20     & 141.8    \\
    \midrule
    \multirow{2}{*}{SSC}            & feedforward   & 700-128-128-35    & 110.8    \\
                                    & recurrent     & 700-128-35    & 110.8    \\
    \bottomrule
  \end{tabular}}
\end{table}

\subsection{Training Configuration}
\label{app: config}
We train the S-MNIST and PS-MNIST datasets for 200 epochs utilizing the Adam optimizer. Their initial learning rates are set to 0.0005 for both feedforward and recurrent networks with the learning rates decaying by a factor of 10 at epochs 60 and 80. For the GSC, SHD, and SSC datasets, we train the models for 100 epochs using the Adam optimizer. 
The initial learning rate of GSC datasets is 0.001 for both feedforward and recurrent networks, which decays by 10 at Epoch 60, 90, and 120.
The initial learning rate is set to 0.0005, and 0.005 for feedforward and recurrent networks on the SHD dataset, with the learning rate decaying to 0.8 times its previous value after every 10 epochs. For the SSC dataset, the initial learning rates are 0.0001 for both feedforward and recurrent networks, and decay to 0.8 times their previous values every 10 epochs. We train S-MNIST, PS-MNIST, and GSC tasks on Nvidia Geforce GTX 3090Ti GPUs with 24GB memory, and train SHD and SSC tasks on Nvidia Geforce GTX 1080Ti GPUs with 12GB memory.


\begin{table}[htb]
  \caption{Infinite norm values with corresponding $\beta_1$ and $\beta_2$ before and after training on S-MNIST.}
  \label{tab: beta_gradient}
  \centering
  \begin{tabular}{ccccc}
    \toprule
    \multicolumn{2}{c}{\textbf{Before}} & \multicolumn{2}{c}{\textbf{After}} & \multicolumn{1}{c}{\multirow{2}{*}{\textbf{Test Acc}}} \\
    $\beta_1$, $\beta_2$         & norm         & $\beta_1$, $\beta_2$      & norm         & \multicolumn{1}{c}{}                    \\
    \midrule
    (-0.2, 0.2) & 1.352 & (-0.184, 0.146) & 1.262 & 98.40 \\
    (-0.2, 0.4) & 1.688 & (-0.188, 0.307) & 1.539 & 99.07 \\
    (-0.2, 0.6) & 2.008 & (-0.203, 0.563) & 1.948 & 99.15 \\
    (-0.2, 0.8) & 2.312 & (-0.202, 0.835) & 2.360 & 89.36 \\
    (-0.4, 0.2) & 1.304 & (-0.379, 0.159) & 1.248 & 99.01 \\
    (-0.4, 0.4) & 1.576 & (-0.370, 0.318) & 1.480 & 99.15 \\
    (-0.4, 0.6) & 1.816 & (-0.383, 0.532) & 1.751 & 99.06 \\
    (-0.4, 0.8) & 2.024 & (-0.380, 0.700) & 1.949 & 99.04 \\
    (-0.6, 0.2) & 1.256 & (-0.621, 0.172) & 1.219 & 99.08 \\
    (-0.6, 0.4) & 1.464 & (-0.621, 0.342) & 1.399 & 99.01 \\
    (-0.6, 0.6) & 1.624 & (-0.634, 0.594) & 1.587 & 98.96 \\
    (-0.6, 0.8) & 1.736 & (-0.612, 0.761) & 1.704 & 98.82 \\
    (-0.8, 0.2) & 1.208 & (-0.812, 0.187) & 1.194 & 98.97 \\
    (-0.8, 0.4) & 1.352 & (-0.815, 0.360) & 1.321 & 99.20 \\
    (-0.8, 0.6) & 1.432 & (-0.812, 0.580) & 1.416 & 98.64 \\
    (-0.8, 0.8) & 1.448 & (-0.821, 0.801) & 1.418 & 98.99 \\      
    \bottomrule
  \end{tabular}
\end{table}

\section{Gradient Exploding Problem Analysis}
We analyze the severity of the gradient exploding problem concerning the TC-LIF initialized in our predefined region. Specifically, recurrent SNNs are trained on the S-MNIST dataset with TC-LIF models initialized by different ($\beta_1$, $\beta_2$) in the second quadrant. Our analysis involves recording the values of $\beta$ before and after training and calculating the infinite norms of the partial derivatives between adjacent time steps in the last hidden layer.

The result reveals that except for the model that is initialized at (-0.2, 0.8) with a convergent infinite norm of 2.36, the remaining models initialized within the second quadrant exhibit commendable performance on the test set.
While an infinite norm of the partial derivative between successive time steps exceeding 1 suggests the potential for the exploding gradient problem during long-term BPTT, our results suggest that values slightly above 1 for the infinite norm do not notably impede convergence. Encouragingly, for the majority of $\beta_1$ and $\beta_2$ initialized in the second quadrant, the corresponding infinite norm values satisfy this condition. Hence, when initializing the TC-LIF model within the second quadrant, stable convergence for the proposed TC-LIF model can be promised, mitigating concerns regarding the gradient exploding problem.

\end{document}